\documentclass[conference]{IEEEtran}
\IEEEoverridecommandlockouts

\usepackage{cite}
\usepackage{tabularx}
\usepackage{amsmath,amssymb,amsfonts}
\usepackage{algorithmic}
\usepackage{graphicx}
\usepackage{textcomp}
\usepackage{xcolor}
\usepackage{booktabs}
\usepackage{subcaption}
\usepackage{placeins}
\usepackage{url}
\usepackage{dblfloatfix}
\usepackage[hidelinks,colorlinks=true, urlcolor=blue]{hyperref}
\usepackage[colorlinks=true, urlcolor=blue]{hyperref}

\usepackage{xcolor}
\usepackage{hyperref}

\hypersetup{
    colorlinks=true,
    linkcolor=black,
    citecolor=black,
    urlcolor=blue
}
\def\BibTeX{{\rm B\kern-.05em{\sc i\kern-.025em b}\kern-.08em
    T\kern-.1667em\lower.7ex\hbox{E}\kern-.125emX}}
\begin{document}
\raggedbottom
\title{Zero-Shot Multimodal Event Spotting in Soccer Broadcasts\\
  }

\author{\IEEEauthorblockN{Sadia Kamal}
\IEEEauthorblockA{\textit{Department of Computer Science and Engineering} \\
\textit{University of Maryland Baltimore County}\\
Maryland, USA \\
sadia1402@umbc.edu}
\and
\IEEEauthorblockN{Mahmudul Hasan}
\IEEEauthorblockA{\textit{Comcast Applied AI} \\
\textit{Comcast}\\
Philadelphia, USA \\
Mdmahmudul\_Hasan@comcast.com}
\and
\IEEEauthorblockN{Tony Braskich}
\IEEEauthorblockA{\textit{Comcast Applied AI} \\
\textit{Comcast}\\
Philadelphia, USA \\
tony\_braskich@comcast.com}
}

\maketitle

\begin{abstract}
Automatic soccer event spotting supports highlight generation, video
indexing, and sports analytics, but existing methods depend on large
annotated datasets and soccer-specific training. We present a zero-shot
framework that detects, classifies, and temporally localizes events in
full-length soccer broadcasts without task-specific parameter training.
Each match is segmented into overlapping clips represented by a compact
visual storyboard and the aligned commentary transcript, which a
general-purpose multimodal model processes jointly. On three manually
annotated matches containing 130 events, the best configuration achieves
72.0\% F1@5s and outperforms SoccerNet-trained CALF and NetVLAD++ in a
cross-dataset setting, while the storyboard representation reduces
inference cost by 72\% compared with individual frames. We further analyze the effects of model family, input representation, and prompt optimization, and introduce an evidence-grounded excitement-scoring stage that ranks detected events with natural-language explanations for highlight selection.
\end{abstract}

\begin{IEEEkeywords}
soccer event detection, zero-shot learning, multimodal large language models, multimodal reasoning.
\end{IEEEkeywords}


\begin{figure*}[t]
    \centering
    \includegraphics[width=0.9\textwidth]{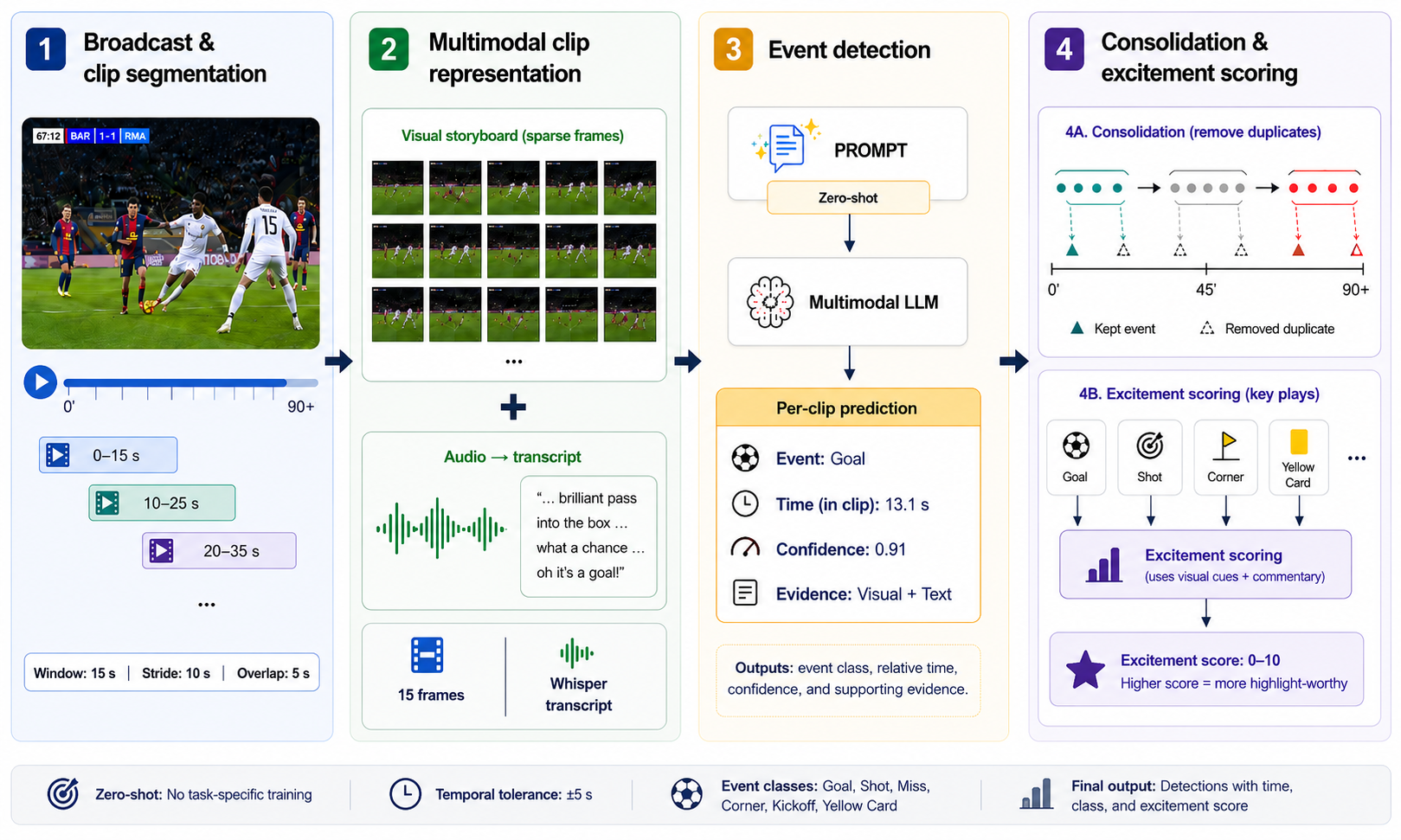}
    \caption{Proposed zero-shot multimodal event-spotting framework. Full-match broadcasts are segmented into overlapping clips represented by sparse visual storyboards and aligned commentary. A multimodal LLM predicts event class, timestamp, confidence, and evidence, followed by temporal consolidation and excitement scoring.}
    \label{fig:pipeline}
\end{figure*}

\section{Introduction}

Soccer broadcasts contain long periods of routine gameplay interrupted by
sparse but important events such as goals, shots, misses, corners, kickoffs,
and disciplinary actions. Automatically detecting and temporally localizing
these events supports applications such as highlight generation, video
indexing, content retrieval, and sports analytics. This task is commonly
formulated as action spotting, where events must be recognized and localized
within complete, untrimmed matches \cite{b1,b2}. It remains challenging because
events are brief, visually similar across classes, and often surrounded by
replays, celebrations, and recap sequences.

Existing soccer action-spotting approaches largely rely on task-specific
models trained on annotated datasets. Prior work has improved performance
through temporal modeling and multimodal information \cite{b3,b4}, while
soccer-specific frameworks such as MatchVision \cite{b5} and SoccerMaster
\cite{b6} use large-scale domain-specific training. These approaches can be
effective but require soccer-specific data, training, and computational
resources.

Multimodal large language models (MLLMs) offer an alternative by jointly
reasoning over visual and textual information without task-specific parameter
fine-tuning. Recent work has also demonstrated their ability to reason about
temporal information in long videos \cite{b7}. However, applying general-purpose
MLLMs to full soccer broadcasts requires detecting sparse events, distinguishing
closely related outcomes, temporally localizing them, and handling replay or
recap content.

In this work, we investigate zero-shot multimodal event spotting and temporal
localization directly from complete soccer broadcasts. We study five questions
covering zero-shot feasibility (\textbf{RQ1}), multimodal input representation
(\textbf{RQ2}), prompt optimization
(\textbf{RQ3}), comparison with supervised soccer-specific methods
(\textbf{RQ4}), and multimodal excitement scoring (\textbf{RQ5}). Full soccer matches are divided into overlapping clips represented using sparse visual information and aligned commentary, from which general-purpose multimodal models jointly predict event categories and their temporal locations.

The main contributions of this work are:

\begin{itemize}
    \setlength{\itemsep}{1pt}
    \item We formulate soccer event spotting over complete broadcasts as a
    zero-shot multimodal task that jointly performs detection,
    classification, and temporal localization without task-specific
    training.
    \item We evaluate the effects of input representation, model choice, and prompt optimization on detection accuracy, temporal localization, and inference cost.
    \item We compare the zero-shot framework with supervised
    action-spotting methods under a common six-class
    evaluation protocol.
    \item We introduce an interpretable excitement-scoring stage for highlight ranking and assess its consistency using two LLM-based judging protocols, including FlowJudge and a stronger GPT-based evaluator.
\end{itemize}

\section{Related Works}

\subsection{Soccer Event Spotting and Multimodal Understanding}

SoccerNet \cite{b1} and SoccerNet-v2 \cite{b2} established benchmarks
for event spotting in soccer broadcasts. Subsequent methods improved
temporal modeling through context-aware learning \cite{b3}, feature
aggregation \cite{b11}, and transformer architectures \cite{b10,b12,b18}.
Multimodal methods incorporate audio and commentary \cite{b4,b17},
while MatchVision \cite{b5} and SoccerMaster \cite{b6} employ
soccer-specific representation learning. X-VARS \cite{b13} explores
multimodal refereeing, and Jiang et al.~\cite{b14} investigate
domain adaptation of vision-language models. In contrast, we study
zero-shot event spotting in full-length broadcasts using general-purpose
multimodal models without task-specific parameter training.

\subsection{Multimodal Temporal Reasoning and Prompt Optimization}

TimeChat \cite{b7}, Grounded-VideoLLM \cite{b15}, and TimeSuite
\cite{b16} improve temporal video understanding through specialized
representations or training. We instead evaluate general-purpose
models for event detection and localization without fine-tuning.
We also investigate MIPROv2 \cite{b9} and GEPA \cite{b8}, which
optimize prompts using task performance and reflection, respectively,
to assess their effects on detection and temporal localization.

\section{Methodology}

\subsection{Dataset}

Our evaluation dataset consists of three full-length soccer broadcasts from publicly available YouTube videos. All three are professionally produced multi-camera broadcast feeds with standard broadcast editing, including camera cuts, close-up views, replays, and retrospective recap sequences, rather than recordings from a single stationary camera. The matches were manually annotated with temporal intervals for six event categories: goal, shot, miss, corner, kickoff, and yellow card, resulting in 130 annotated events. The source video URLs, temporal annotations, and evaluation metadata are publicly available in our \href{https://github.com/sadia14/zero-shot-soccer-event-spotting-IEEE-ISM_2026}{\textcolor{blue}{GitHub repository}}.

Each match is segmented into 15-second clips with a 5-second overlap, with 15 frames uniformly sampled per clip. Commentary is transcribed using Whisper Medium and aligned with each clip. We use this configuration as the default after preliminary experiments with shorter, denser clips produced higher recall but lower precision.

\subsection{Problem Formulation}
A match $V$ is divided into overlapping clips
$\mathcal{C}=\{C_1,\ldots,C_N\}$, where each clip
$C_i=(S_i,\mathcal{T}_i,t_i^{\mathrm{start}},t_i^{\mathrm{end}})$ consists of
a storyboard $S_i$ of sampled frames, the aligned commentary transcript
$\mathcal{T}_i$, and its start and end times on the match timeline. For each
clip, the model outputs
$\hat{y}_i=(\hat{d}_i,\hat{e}_i,\hat{\tau}_i,\hat{c}_i,\hat{r}_i)$: a
detection flag, an event category, a clip-relative timestamp
$\hat{\tau}_i\in[0,\,t_i^{\mathrm{end}}-t_i^{\mathrm{start}}]$, a confidence
score, and supporting evidence. Detected events are mapped to the match
timeline as
\begin{equation}
\hat{t}_i^{\mathrm{abs}} = t_i^{\mathrm{start}} + \hat{\tau}_i.
\label{eq:absolute_time}
\end{equation}
Unlike event-centered classification, the model must jointly decide whether
an event occurs, what it is, and when it occurs.
 \subsection{Pipeline Overview}

As illustrated in Fig.~\ref{fig:pipeline}, the framework processes overlapping
clips using sparse frame sampling, storyboard construction, commentary
alignment, multimodal event detection, temporal localization, consolidation,
and excitement scoring.

\subsubsection{Sparse Frame Sampling}
Frames are uniformly sampled across each clip to provide observations throughout
the temporal window. The default configuration extracts 15 frames from each
15-second clip.

\subsubsection{Storyboard Construction}

Sampled frames are resized to $384 \times 216$ pixels and arranged
chronologically into a single storyboard, with each frame associated with its
relative timestamp. For the default 15-frame configuration, the frames are
organized in a $5 \times 3$ grid, resulting in a storyboard resolution of
$1920 \times 648$ pixels. This compact representation allows the model to
reason over temporal progression within a single visual input while reducing
the number of separate visual inputs. We compare this representation with
providing the same sampled frames individually at the same per-frame
resolution.

\subsubsection{Commentary Alignment}
The aligned commentary transcript is provided alongside the visual input,
providing complementary semantic evidence about event occurrence and outcome.

\subsubsection{Multimodal Event Detection}
The visual representation and aligned commentary are jointly provided to a
general-purpose multimodal model. The model predicts one event category or
\textit{no important event}, together with a relative timestamp, confidence
score, and supporting visual and textual evidence. Outputs follow a structured
format for automatic parsing and evaluation.





\subsubsection{Absolute Temporal Localization}

Each predicted clip-relative timestamp $\hat{\tau}_i$ is mapped to the full-match
timeline using Eq.~\ref{eq:absolute_time}, allowing predictions from different
clips to be compared on a common temporal axis.

\subsubsection{Temporal Consolidation}

Subsequent overlapping clips can produce multiple predictions for the same event. Predictions are grouped by event category and ordered by absolute timestamp. Same-class detections are clustered using a 20-second temporal window, with a 30-second window for goals and corners, and the highest-confidence prediction in each cluster is retained. Predictions from different event classes are never merged.

\subsection{Prompt Optimization}
\label{sec:prompt_optimization}

We compare manually designed and automatically optimized prompts while keeping
model parameters fixed.
\paragraph{Manual storyboard prompt}
Our primary prompt is designed for the storyboard representation. It defines
the event taxonomy, provides event-specific decision guidance, and instructs
the model to reason jointly over the temporally ordered visual evidence and
aligned commentary. The model considers the progression of the clip before
assigning an event label and is instructed to distinguish live events from
replay or recap segments. Predictions follow a structured format containing
the event decision, category, relative timestamp, confidence score, and
supporting visual and textual evidence.

\paragraph{Two-prompt decomposition}
We additionally evaluate a two-prompt configuration that separates recognition
from verification. The first prompt performs event recognition and coarse
temporal localization, while the second re-examines the initial prediction and
multimodal evidence to produce the final event decision and refined timestamp.
This tests whether separating initial recognition from verification improves
prediction reliability.

\paragraph{Automatic prompt optimization} We additionally apply GEPA and MIPROv2 to investigate whether the manually designed instructions can be systematically improved. Both methods use labeled development examples to optimize task-specific instructions. Separate training, validation, and held-out test splits are used while all model parameters remain fixed. The optimization modifies the textual instructions and reasoning guidance while preserving the event taxonomy, multimodal input representation, and structured output format. The optimized prompts are fixed and applied unchanged during full-match inference. Thus, GEPA and MIPROv2 perform label-guided prompt optimization without task-specific parameter fine-tuning, allowing us to examine how prompt structure affects detection and temporal localization within the same parameter-frozen inference framework.

\subsection{Models}

We evaluate several models under the same zero-shot soccer event-detection
setting, including GPT-5.5, GPT-5.6 Sol, GPT-5.6 Terra, GPT-5.6 Luna,
Qwen3.5-4B, and Qwen3-VL-8B Embedding. The GPT variants are selected to
compare different capability--cost operating points within the same model
family, while Qwen3.5-4B provides a lightweight open-source generative
baseline that can be deployed on a single consumer GPU. The generative models receive either sampled frames or storyboards
together with aligned commentary and produce the event category, relative
timestamp, confidence score, and supporting multimodal evidence.

We additionally evaluate Qwen3-VL-8B Embedding as an embedding-based baseline,
where clip representations are compared with event-description prototypes.
This allows us to examine whether generative multimodal reasoning provides an
advantage over representation-based similarity matching. All models are
evaluated without task-specific fine-tuning under the same evaluation protocol.

\subsection{Excitement Scoring}


For excitement analysis, we evaluate the 175 post-consolidation predictions belonging to the six event categories used in the main event-spotting evaluation. Predictions outside the benchmark taxonomy are excluded to maintain consistency with the detection evaluation. Each retained event is scored once, avoiding repeated assessments of the same event arising from overlapping clips. GPT-5.4 Mini scores each event using its predicted category, detection confidence, supporting visual evidence, and aligned transcript rather than reprocessing the original video.

\begin{equation}
S_{\mathrm{exc}}
=
w_1 S_{\mathrm{event}}
+
w_2 S_{\mathrm{visual}}
+
w_3 S_{\mathrm{text}},
\end{equation}

where $w_1$, $w_2$, and $w_3$ control the relative contribution of event
importance, visual intensity, and transcript intensity, respectively, with
$w_1+w_2+w_3=1$.

In our experiments, we set $w_1=0.40$, $w_2=0.35$, and $w_3=0.25$.
These values place greater emphasis on event importance and visual evidence
while retaining commentary as complementary semantic and emotional context.
The weights are heuristic design choices rather than learned parameters and can
be adjusted depending on the desired highlight-selection behavior or
application requirements.

In addition to the excitement score, the model generates a confidence score
and a concise natural-language explanation describing the evidence supporting
its assessment, providing an interpretable estimate of event excitement for
highlight ranking.

\section{Evaluation}
\subsection{Evaluation Protocol}

We evaluate event detection using event-level precision, recall, F1 score,
mean Average Precision (mAP), and temporal localization error. Since predictions
are single timestamps while ground-truth events are temporal intervals, a
prediction is temporally matched if it lies within the annotated interval or
within a predefined tolerance around it.

Unless otherwise stated, we use a temporal tolerance of $\pm5$ seconds. A
prediction is a true positive (TP) when it matches a ground-truth event of the
same category within this tolerance. Unmatched predictions are false positives
(FP), and unmatched ground-truth events are false negatives (FN). Each
prediction and ground-truth event can participate in at most one match. The 130 annotated events constitute the ground-truth set for the six-class event-spotting evaluation. Prediction counts are computed after temporal consolidation and therefore need not equal the ground-truth count; unmatched predictions are counted as false positives. 

We report Precision@5s, Recall@5s, and F1@5s. For matched events,
temporal localization is measured by the mean absolute error (MAE)
between each prediction and its ground-truth interval, taken as zero
inside the interval and as the distance to the nearest boundary otherwise. For mAP, predictions for each event category are ranked by confidence and
average precision (AP) is computed from the resulting precision--recall curve.
mAP is obtained by averaging AP across event categories using the same
$\pm5$-second temporal matching criterion.




\subsection{Excitement-Score Evaluation}
 \label{sec:llm-judge}

We assess excitement-score consistency over the same 175 post-consolidation event predictions using two LLM-based judges: FlowJudge \cite{b20} and GPT-5.6 Sol. Each judge receives the predicted event category, extracted visual evidence, and aligned commentary, while the original GPT-5.4 Mini excitement score, confidence, and explanation are withheld to avoid anchoring. Both judges assign scores on the same 0--10 scale. Because the intended application is event ranking, Spearman's rank correlation ($\rho$) \cite{b19}  is the primary agreement metric; we additionally report mean absolute error (MAE) to characterize differences in absolute score calibration.

\section{Results}

All experiments were conducted using AWS EC2 \texttt{g5.2xlarge} instances as
the common experimental environment. Preprocessing, feature extraction,
baseline execution, and open-source model inference were performed locally,
while API-based multimodal models were accessed through their respective APIs.
Results are organized around the five research questions introduced earlier.

\subsection{RQ1: Overall Event Detection Performance}
\begin{table*}[!t]
\centering
\caption{Average event-detection performance across three matches, with task-specific training and per-match inference cost.}
\label{tab:overall_results}

\resizebox{\textwidth}{!}{
\begin{tabular}{lllccccccc}
\toprule
\textbf{Model} &
\textbf{Training} &
\textbf{Input} &
\textbf{P@5s} &
\textbf{R@5s} &
\textbf{F1@5s} &
\textbf{mAP@5s} &
\textbf{MAE (s)} &
\textbf{Pred./match} &
\textbf{Cost/match} \\
\midrule

CALF
& Yes (SoccerNet)
& Video
& 12.91
& 43.06
& 19.60
& 31.95
& 0.94
& 147
& N/A \\

NetVLAD++
& Yes (SoccerNet-v2)
& Video
& 27.23
& 65.96
& 38.49
& 56.82
& 0.81
& 103
& N/A \\

\midrule

GPT-5.5
& No
& Storyboard
& \textbf{63.85}
& \textbf{82.61}
& \textbf{72.02}
& \textbf{76.24}
& \textbf{0.27}
& 56
& \$18.01 \\

GPT-5.6 Sol
& No
& Storyboard
& 60.93
& 82.45
& 70.05
& 74.77
& 0.47
& 58
& \$18.01 \\

GPT-5.6 Terra
& No
& Storyboard
& 59.90
& 77.98
& 67.75
& 74.08
& 0.35
& 56
& \$7.20 \\

GPT-5.6 Luna
& No
& Storyboard
& 45.83
& 77.19
& 57.32
& 67.16
& 0.30
& 74
& \$0.72 \\

Qwen3.5-4B
& No
& Storyboard
& 30.46
& 60.59
& 39.73
& 17.96
& 1.14
& 88
& \$0.25 \\

Qwen3-VL-8B
& No
& Storyboard
& 17.17
& 19.38
& 16.96
& 34.27
& 0.37
& 56
& \$0.25 \\

\bottomrule
\end{tabular}
}
\end{table*}

\begin{table*}[t]
\centering
\caption{Class-wise event recall across the three soccer matches. Each entry
reports correctly detected over annotated events for each class.}
\label{tab:class_recall}
\footnotesize
\begin{tabular*}{\textwidth}{@{\extracolsep{\fill}}lcccccc@{}}
\toprule
\textbf{Model} &
\textbf{Corner} &
\textbf{Goal} &
\textbf{Kickoff} &
\textbf{Miss} &
\textbf{Shot} &
\textbf{Yellow Card} \\
\midrule
GPT-5.5       & 16/22 (72.7\%) & 18/21 (85.7\%) & \textbf{6/6 (100.0\%)} & 19/26 (73.1\%) & 35/41 (85.4\%) & \textbf{14/14 (100.0\%)} \\
GPT-5.6 Sol   & 16/22 (72.7\%) & 17/21 (81.0\%) & \textbf{6/6 (100.0\%)} & 18/26 (69.2\%) & \textbf{37/41 (90.2\%)} & \textbf{14/14 (100.0\%)} \\
GPT-5.6 Terra & 17/22 (77.3\%) & 17/21 (81.0\%) & \textbf{6/6 (100.0\%)} & 17/26 (65.4\%) & 31/41 (75.6\%) & \textbf{14/14 (100.0\%)} \\
GPT-5.6 Luna  & 15/22 (68.2\%) & 18/21 (85.7\%) & \textbf{6/6 (100.0\%)} & 15/26 (57.7\%) & 33/41 (80.5\%) & 12/14 (85.7\%) \\
Qwen3.5-4B    & 14/22 (63.6\%) & \textbf{19/21 (90.5\%)} & 3/6 (50.0\%) & 15/26 (57.7\%) & 14/41 (34.1\%) & \textbf{14/14 (100.0\%)} \\
Qwen3-VL-8B   & 7/22 (31.8\%)  & 11/21 (52.4\%) & 0/6 (0.0\%) & 0/26 (0.0\%) & 0/41 (0.0\%) & 7/14 (50.0\%) \\
\midrule
CALF          & 18/22 (81.8\%) & 8/21 (38.1\%)  & 1/6 (16.7\%) & 9/26 (34.6\%) & 12/41 (29.3\%) & 8/14 (57.1\%) \\
NetVLAD++     & \textbf{21/22 (95.5\%)} & 10/21 (47.6\%) & 3/6 (50.0\%) & \textbf{20/26 (76.9\%)} & 20/41 (48.8\%) & 11/14 (78.6\%) \\
\bottomrule
\end{tabular*}
\end{table*}

\begin{figure}[t]
    \centering
    \includegraphics[width=0.8\linewidth]{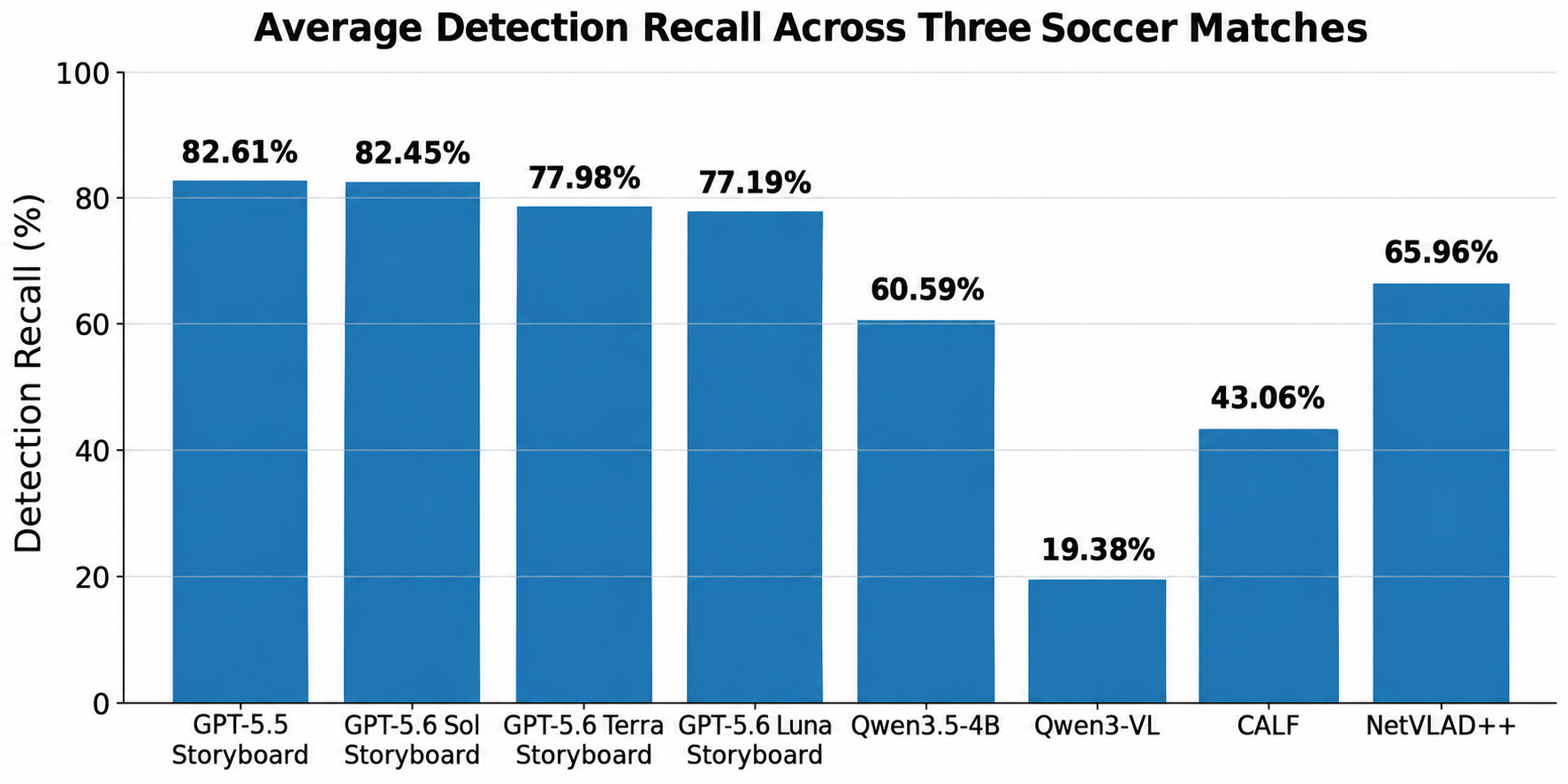}
    \caption{Average detection recall across three matches.
    GPT-5.5 and GPT-5.6 Sol achieve the highest recall, while
    NetVLAD++ and CALF obtain 65.96\% and 43.06\%, respectively.}
    \label{fig:detection_recall}
\end{figure}
Table~\ref{tab:overall_results} summarizes average event-detection performance
across the three matches. Among the evaluated multimodal configurations,
GPT-5.5 with storyboard input achieves the highest overall performance, with 63.85\% Precision@5s, 82.61\% Recall@5s, 72.02\% F1@5s, and 76.24\% mAP@5s.
It also achieves the lowest temporal localization error of 0.27 seconds.

GPT-5.6 Sol performs comparably, achieving 82.45\% Recall@5s and 70.05\%
F1@5s. The 0.16-point recall difference is small in this three-match evaluation. GPT-5.6 Terra provides a lower-cost alternative, achieving 67.75\% F1@5s at
\$7.20 per match, compared with \$18.01 for GPT-5.5 and GPT-5.6 Sol.
GPT-5.6 Luna further reduces the cost to \$0.72 while retaining 77.19\%
recall, but its lower precision (45.83\%) reduces F1@5s to 57.32\%.

The Qwen models perform substantially worse overall. Qwen3.5-4B achieves
60.59\% recall but 30.46\% precision, while Qwen3-VL-8B obtains only
19.38\% recall. Within our three-match evaluation, these results show substantial variation in zero-shot event-spotting capability across multimodal model families.

Fig.~\ref{fig:detection_recall} compares Recall@5s across all methods: the
GPT configurations recover 77--83\% of annotated events, compared with
65.96\% for NetVLAD++ and 43.06\% for CALF.

Table~\ref{tab:class_recall} further shows substantial variation across event
categories. Qwen3.5-4B obtains the highest goal recall (90.5\%) while detecting
only 34.1\% of shots, and NetVLAD++ achieves the highest corner recall
(95.5\%) despite lower aggregate performance. In contrast, the stronger GPT
configurations provide more balanced recall across event categories.
Qwen3-VL-8B fails to recover kickoffs, misses, and shots.

Table~\ref{tab:overall_results} reports metrics macro-averaged across matches,
whereas Table~\ref{tab:class_recall} pools detections across matches, explaining
the small differences between aggregate recall and pooled class-wise counts.
Overall, \textbf{RQ1} shows that, on the evaluated broadcasts, general-purpose multimodal models can perform soccer event spotting and temporal localization without soccer-specific parameter fine-tuning, although performance varies across model families.

\subsection{RQ2: Effect of Input Representation and Modality}

To isolate the effect of input representation and modality, we conduct this
ablation on a single full-length match using GPT-5.6. We compare four
configurations: transcript only, 15 visual frames only, 15 frames with
transcript, and storyboard with transcript, using the same event taxonomy,
evaluation protocol, and post-processing.

Table~\ref{tab:input_ablation} shows that transcript-only input achieves
50.00\% F1@5s, while visual frames alone improve it to 63.57\%. Combining
frames with commentary further increases F1@5s to 72.31\%, demonstrating the
benefit of multimodal information. The storyboard achieves the highest
Recall@5s (85.00\%) and F1@5s (73.91\%). Compared with individual frames plus
transcript, it also reduces inference cost from \$64.69 to \$18.01 (72.2\%)
while slightly improving F1@5s.

\begin{table}[t]
\centering
\caption{Input representation and modality ablation on a single full-length
soccer match using GPT-5.6 with a $\pm5$-second matching tolerance.}
\label{tab:input_ablation}

\footnotesize
\setlength{\tabcolsep}{2pt}
\renewcommand{\arraystretch}{0.9}

\begin{tabular}{lcccccc}
\toprule
\textbf{Input} &
\textbf{P@5s} &
\textbf{R@5s} &
\textbf{F1@5s} &
\textbf{MAE} &
\textbf{Pred.} &
\textbf{Cost} \\
\midrule
Transcript only         & 53.85 & 46.67 & 50.00 & 0.68 & 52 & \textbf{13.56} \\
15 frames only          & 59.42 & 68.33 & 63.57 & \textbf{0.24} & 69 & 64.47 \\
15 frames + transcript  & \textbf{67.14} & 78.33 & 72.31 & 0.91 & 70 & 64.69 \\
Storyboard + transcript & 65.38 & \textbf{85.00} & \textbf{73.91} & 0.43 & 78 & 18.01 \\
\bottomrule
\end{tabular}

\end{table}

Class-wise, commentary alone performs poorly on outcome-sensitive events such
as misses (22.2\%), while combining visual and textual evidence raises miss
recall to 77.8\%, indicating that the modalities are complementary. Overall,
\textbf{RQ2} indicates that multimodal input improves event spotting and that,
in this single-match ablation, the storyboard gives the strongest aggregate
performance at substantially lower cost.

\subsection{RQ3: Prompt Structure and Optimization}

\begin{table*}[t]
\centering
\caption{Comparison of manual and automatically optimized prompting strategies
across three soccer matches.}
\label{tab:prompt_optimization}

\small
\setlength{\tabcolsep}{5pt}
\begin{tabular}{llcccccc}
\toprule
\textbf{Prompt Strategy} &
\textbf{Model} &
\textbf{P@5s} &
\textbf{R@5s} &
\textbf{F1@5s} &
\textbf{MAE (s)} &
\textbf{Pred./match} &
\textbf{Cost (\$)} \\
\midrule

Storyboard
& GPT-5.6 Sol
& 60.93\%
& \textbf{82.45\%}
& \textbf{70.05\%}
& 0.47
& 58
& 18.01 \\

Storyboard-2
& GPT-5.6 Sol
& \textbf{62.24\%}
& 71.5\%
& 61.30\%
& \textbf{0.14}
& 54
& 18.01 \\

GEPA
& GPT-5.6 Sol
& 55.41\%
& 71.83\%
& 62.17\%
& 0.23
& 56
& 20.00 \\

MIPROv2
& GPT-5.6 Sol
& 53.09\%
& 80.37\%
& 63.89\%
& 0.27
& 65
& 20.00 \\

\bottomrule
\end{tabular}
\end{table*}

Table~\ref{tab:prompt_optimization} compares the manually designed storyboard
prompt, a two-prompt variant, and prompts optimized using GEPA and MIPROv2.
Here, \textit{manually designed} denotes a researcher-authored prompt specifying
the event taxonomy, decision rules, temporal reasoning guidance, and structured
output format without automated optimization. The manually designed storyboard prompt achieves the highest Recall@5s
(82.45\%) and F1@5s (70.05\%). Storyboard-2 produces a more conservative
detector, with the highest precision (62.24\%), fewest predictions (54), and
lowest temporal MAE (0.14 s), but its recall decreases to 71.5\%.

Among the automatically optimized prompts, MIPROv2 achieves 80.37\% recall,
approaching the storyboard prompt, but its lower precision (53.09\%) results
in 63.89\% F1@5s. GEPA achieves 55.41\% precision, 71.83\% recall, and
62.17\% F1@5s. Both optimized prompts improve temporal localization over the
storyboard baseline, with MAEs of 0.23 s for GEPA and 0.27 s for MIPROv2
compared with 0.47 s for the storyboard prompt.



\begin{table*}[t]
\centering
\caption{Event-level Recall@5s across event classes for different prompt strategies.}
\label{tab:prompt_class_results}

\small
\setlength{\tabcolsep}{5pt}
\begin{tabular}{llcccccc}
\toprule
\textbf{Prompt} &
\textbf{Model} &
\textbf{Corner} &
\textbf{Goal} &
\textbf{Kickoff} &
\textbf{Miss} &
\textbf{Shot} &
\textbf{Yellow Card} \\
\midrule

Storyboard
& GPT-5.6 Sol
& \textbf{72.7\% (16/22)}
& \textbf{81.0\% (17/21)}
& \textbf{100.0\% (6/6)}
& 69.2\% (18/26)
& \textbf{90.2\% (37/41)}
& \textbf{100.0\% (14/14)} \\

Storyboard-2
& GPT-5.6 Sol
& 63.6\% (14/22)
& 76.2\% (16/21)
& 66.7\% (4/6)
& 65.4\% (17/26)
& 73.2\% (30/41)
& 85.7\% (12/14) \\

GEPA
& GPT-5.6 Sol
& \textbf{72.7\% (16/22)}
& \textbf{81.0\% (17/21)}
& 83.3\% (5/6)
& 65.4\% (17/26)
& 68.3\% (28/41)
& 85.7\% (12/14) \\

MIPROv2
& GPT-5.6 Sol
& 68.2\% (15/22)
& \textbf{81.0\% (17/21)}
& \textbf{100.0\% (6/6)}
& \textbf{73.1\% (19/26)}
& 85.4\% (35/41)
& \textbf{100.0\% (14/14)} \\

\bottomrule
\end{tabular}
\end{table*}

Table~\ref{tab:prompt_class_results} shows that the effect of prompt
optimization is event dependent. The storyboard prompt achieves the highest
shot recall (90.2\%) and ties for the highest recall on several other classes.
MIPROv2 achieves the highest miss recall (73.1\%) while matching the storyboard
on goals, kickoffs, and yellow cards. GEPA matches storyboard recall for
corners and goals but performs worse on the remaining classes, while
Storyboard-2 generally exhibits lower class-wise recall.

Although GEPA and MIPROv2 are designed to improve task-specific prompting,
their gains are limited in our setting because many errors arise from the
multimodal evidence itself rather than from the textual instructions alone.
Sparse visual sampling, ambiguous event outcomes, and broadcast or recap content
can leave insufficient evidence for distinguishing closely related events.
Prompt optimization can therefore modify the detector's decision behavior,
but it cannot recover visual or temporal information that is absent from the
input. This helps explain why the optimized prompts improve some classes or
temporal localization without consistently improving overall F1.

Overall, \textbf{RQ3} shows that automatic prompt optimization does not uniformly
improve event spotting but changes the detector's precision, recall, temporal
localization, and class sensitivity. MIPROv2 favors broader event coverage,
while GEPA improves temporal localization but has lower aggregate detection
performance. The manually designed storyboard prompt achieves the highest
Recall@5s and F1@5s among the evaluated strategies.






\subsection{RQ4: Comparison with Soccer-Specific Baselines}

We compare the proposed zero-shot multimodal approach with CALF and NetVLAD++,
two supervised action-spotting models pretrained on SoccerNet. Both baselines
are evaluated on the same three matches using the same six-class taxonomy and
temporal evaluation protocol.

As shown in Table~\ref{tab:overall_results}, NetVLAD++ achieves 65.96\%
Recall@5s and 38.49\% F1@5s, while CALF achieves 43.06\% Recall@5s and
19.60\% F1@5s. Both methods also produce substantially more predictions than
the strongest storyboard-based configurations.

Under this common six-class evaluation, the strongest zero-shot multimodal
configurations achieve higher precision, recall, F1@5s, and mAP@5s than the
evaluated supervised baselines. However, CALF and NetVLAD++ are pretrained on
SoccerNet and evaluated here without retraining. Their results therefore
represent cross-dataset transfer rather than their original in-domain
performance, and differences in annotation conventions and broadcast
characteristics may affect the comparison.

Overall, \textbf{RQ4} shows that general-purpose multimodal models can achieve
strong cross-dataset soccer event-spotting performance without soccer-specific
parameter fine-tuning.

\subsection{RQ5: Excitement-Score Analysis}
\label{sec:excitement_results}



\begin{figure*}[t]
    \centering
    \begin{subfigure}[b]{0.30\textwidth}
        \centering
        \includegraphics[width=\linewidth]{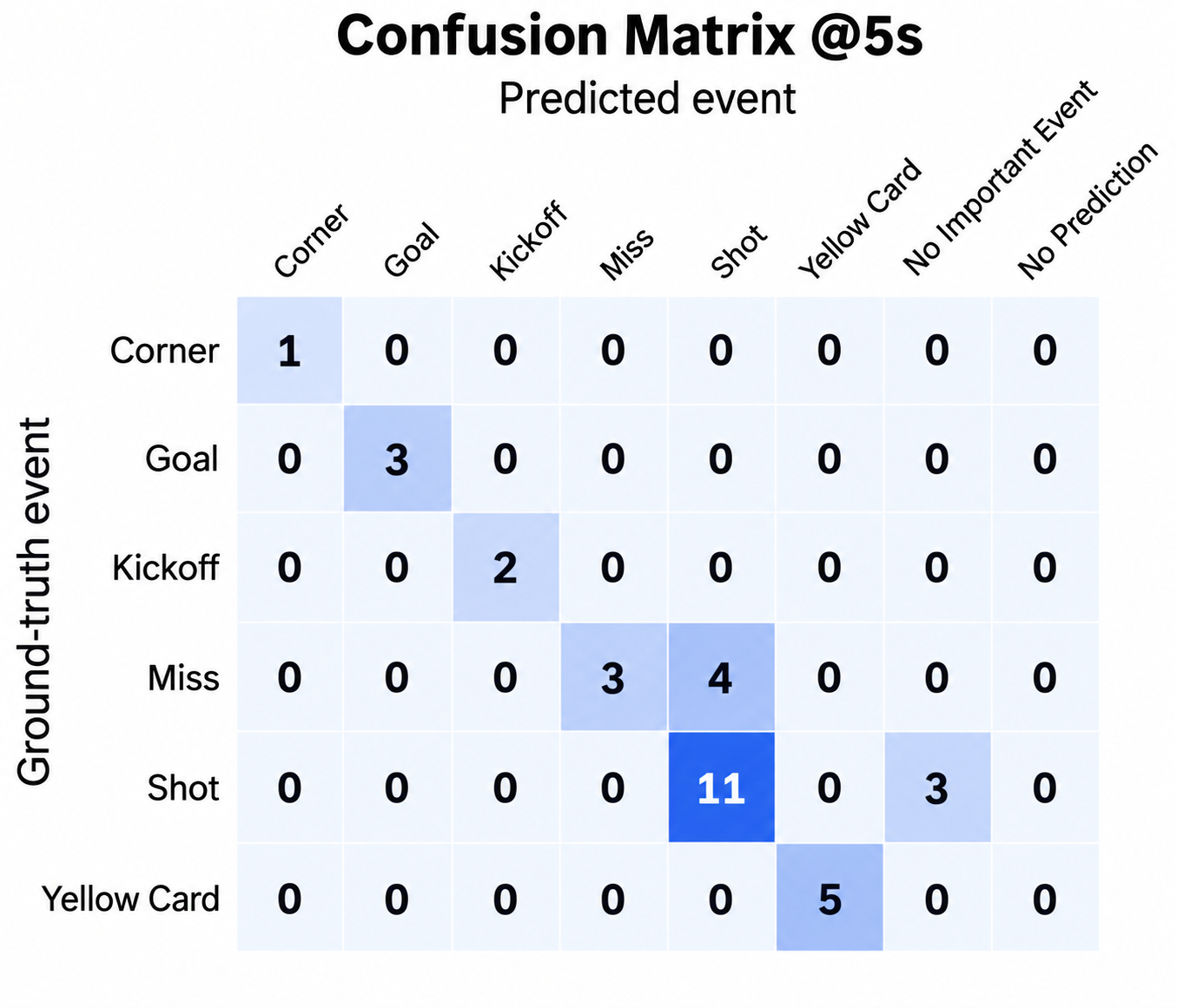}
        \caption{Match 1}
        \label{fig:confusion_a}
    \end{subfigure}
    \hfill
    \begin{subfigure}[b]{0.30\textwidth}
        \centering
        \includegraphics[width=\linewidth]{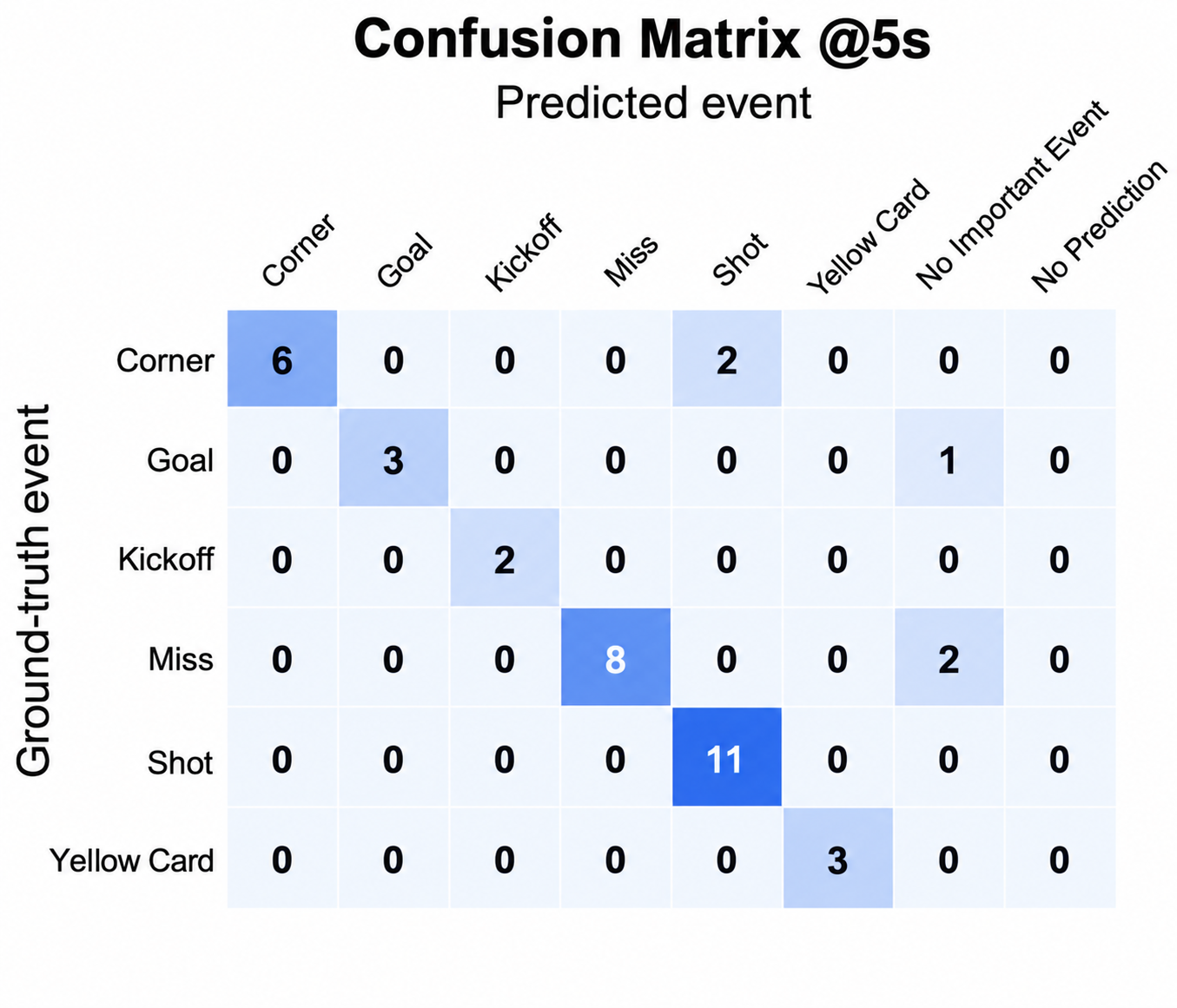}
        \caption{Match 2}
        \label{fig:confusion_b}
    \end{subfigure}
    \hfill
    \begin{subfigure}[b]{0.30\textwidth}
        \centering
        \includegraphics[width=\linewidth]{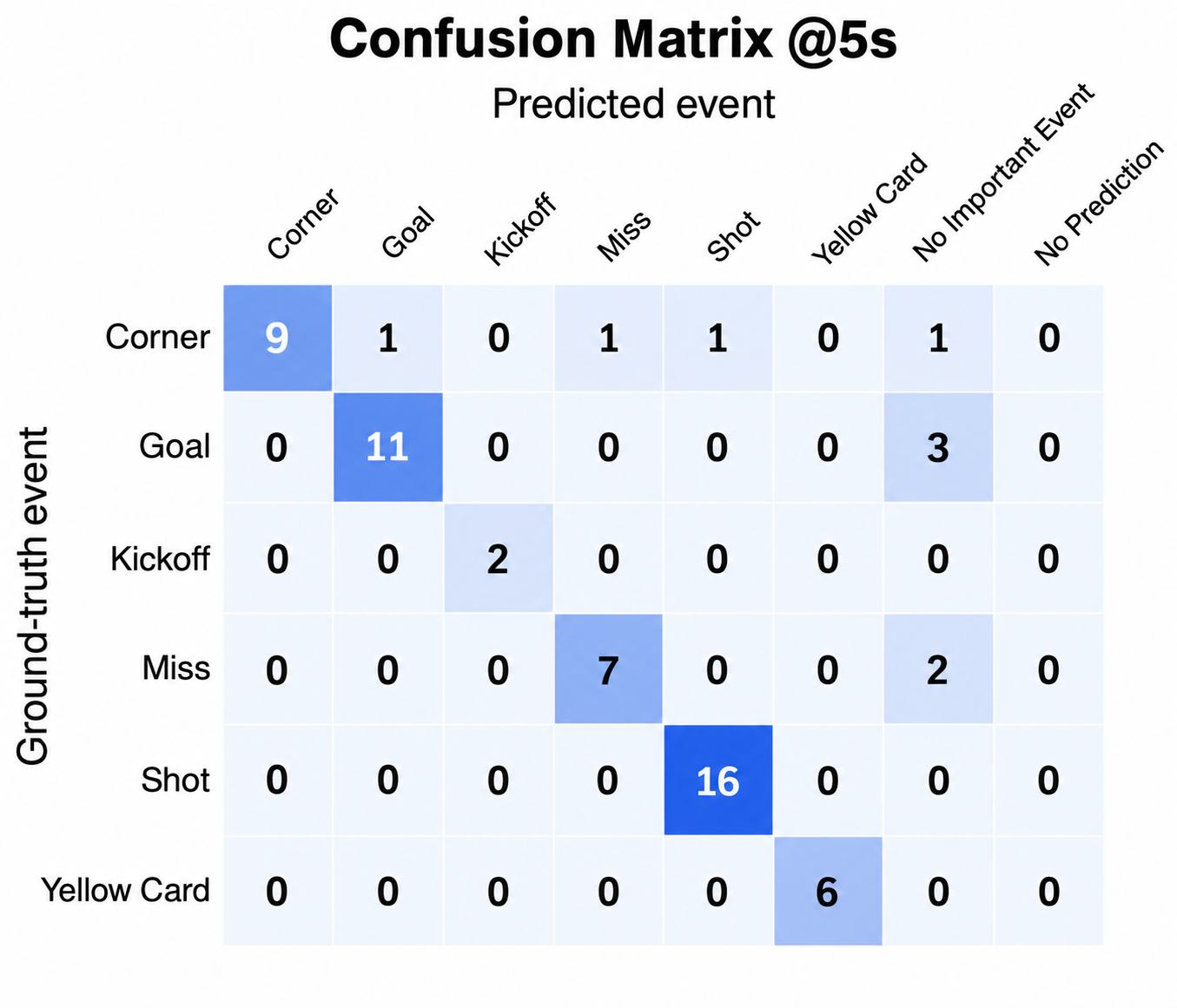}
        \caption{Match 3}
        \label{fig:confusion_c}
    \end{subfigure}

    \caption{Per-match event-level confusion matrices for GPT-5.6 Sol with
    storyboard and transcript input. Rows denote ground-truth classes and
    columns predicted classes.}
    \label{fig:error_confusion}
\end{figure*}
We evaluate excitement scoring over 175 post-consolidation event predictions across the three matches and six event categories. Unlike the clip-level detector outputs, each retained prediction represents a single event candidate after temporal consolidation, preventing overlapping detections of the same event from being counted repeatedly.

As shown in Table~\ref{tab:excitement_class}, GPT-5.4 Mini assigns an overall mean excitement score of 5.07. Goals receive the highest average score (7.87), followed by shots (5.81) and misses (5.03), while corners (2.66), kickoffs (2.44), and yellow cards (2.36) receive lower scores. This ordering is broadly preserved by both LLM-based judges, indicating that the scoring framework differentiates events according to estimated highlight relevance.

To assess scoring consistency, we compare GPT-5.4 Mini with FlowJudge and a stronger GPT-based judge. As shown in Table VII, FlowJudge achieves an overall Spearman rank correlation of $\rho=0.716$ with an MAE of 1.37, while the GPT-based judge achieves substantially stronger agreement with $\rho=0.926$ and an MAE of 0.64. FlowJudge assigns a lower mean score than GPT-5.4 Mini (4.46 versus 5.07), indicating more conservative absolute-score calibration, whereas the GPT-based judge produces a similar overall mean score (5.25). Positive rank agreement is observed across all three matches for both judges.

Overall, RQ5 demonstrates consistent excitement ranking across two LLM-based judges, while human evaluation remains necessary to establish alignment with perceived highlight relevance.
\begin{table}[t]
\centering
\caption{Excitement-score analysis over 175 post-consolidation event predictions from the six benchmark categories across the three evaluated matches.}
\label{tab:excitement_class}

\small
\setlength{\tabcolsep}{2.5 pt}
\renewcommand{\arraystretch}{1}

\begin{tabular}{lccccc}
\hline
\textbf{Event} & \textbf{N} & \textbf{GPT Avg.} & \textbf{Avg. Conf.} & \textbf{Flow Avg.} & \textbf{GPT-J Avg.} \\
\hline
Corner      & 21  & 2.66 & 0.79 & 2.67 & 3.38 \\
Goal        & 27  & 7.87 & 0.92 & 7.70 & 8.00 \\
Kickoff     & 11  & 2.44 & 0.86 & 1.45 & 2.18 \\
Miss        & 26  & 5.03 & 0.85 & 4.46 & 4.81 \\
Shot        & 72  & 5.81 & 0.82 & 4.69 & 5.96 \\
Yellow Card & 18  & 2.36 & 0.87 & 2.56 & 3.00 \\
\hline
\textbf{Overall} & \textbf{175} & \textbf{5.07} & \textbf{0.84} & \textbf{4.46} & \textbf{5.25} \\
\hline
\end{tabular}
\end{table}
\begin{table*}[t]
\centering
\caption{Agreement of GPT-5.4 Mini excitement scores with two LLM-based judges.}
\label{tab:judge_agreement}
\resizebox{\textwidth}{!}{%
\begin{tabular}{lcccccccc}
\hline
\textbf{Match} &
\textbf{N} &
\textbf{GPT Avg.} &
\textbf{Flow Avg.} &
\textbf{Flow $\rho$} &
\textbf{Flow MAE} &
\textbf{GPT-Judge Avg.} &
\textbf{GPT-Judge $\rho$} &
\textbf{GPT-Judge MAE} \\
\hline
Match 1 & 78 & 5.05 & 4.46 & 0.785 & 1.28 & 5.28 & 0.943 & 0.57 \\
Match 2  & 43 & 5.52 & 4.53 & 0.683 & 1.51 & 5.67 & 0.908 & 0.66 \\
Match 3    & 54 & 4.74 & 4.39 & 0.581 & 1.38 & 4.87 & 0.880 & 0.72 \\
\hline
\textbf{Overall} &
\textbf{175} &
\textbf{5.07} &
\textbf{4.46} &
\textbf{0.716} &
\textbf{1.37} &
\textbf{5.25} &
\textbf{0.926} &
\textbf{0.64} \\
\hline
\end{tabular}%
}
\end{table*}

\subsection{Error Analysis}
\label{sec:error_analysis}

\paragraph{Class ambiguity}
We examine event-level confusion matrices for the GPT-5.6 Sol storyboard
configuration across the three matches using the same $\pm5$\,s temporal
tolerance as the main evaluation.

The confusion patterns show that errors are concentrated among semantically
related attacking events. In Match~1, four of seven misses are classified as
shots, indicating difficulty distinguishing an off-target miss from an
on-target attempt. In Matches~2 and~3, corner errors sometimes correspond to
the goal-directed action following the set piece. Event rejection also
contributes to errors, including three of fourteen shots in Match~1 and three
of fourteen goals in Match~3 classified as \textit{No Important Event}.

These patterns suggest that class ambiguity depends on the temporal context
within a clip. A corner may be immediately followed by a shot, while
distinguishing a shot from a miss requires evidence of the attempt's final
trajectory. Sparse storyboard sampling may therefore capture the attacking
action while providing less explicit evidence of its final outcome.

\paragraph{Broadcast--recap confusion}
We additionally conduct a qualitative analysis of broadcast state for goal
predictions. An event is labeled \textit{broadcast} when the clip contains its
original occurrence and \textit{recap} when a later segment revisits an earlier
event. Broadcast state is determined manually by inspecting the surrounding
temporal context.

Among four manually inspected goal predictions from one representative match,
three correspond to the original broadcast occurrence and one to a recap. In
the recap case, a previously occurring scoring sequence is shown again and
interpreted as a new goal. Although this small qualitative sample does not
support estimating a general recap-confusion rate, it demonstrates that
recap content can produce false-positive event detections despite explicit
prompt instructions to distinguish live events from retrospective footage.

These observations suggest that prompt-based broadcast-state reasoning can
handle some replay or recap sequences but is not fully reliable. Explicit
broadcast-state modeling may provide a more robust mechanism for distinguishing
new events from retrospective content.

\begin{table*}[t]
\centering
\caption{Representative excitement-scoring examples from the three matches.}
\label{tab:excitement_examples}

\small
\renewcommand{\arraystretch}{0.95}
\setlength{\tabcolsep}{3pt}
\setlength{\textfloatsep}{5pt}
\setlength{\floatsep}{4pt}
\setlength{\intextsep}{4pt}
\renewcommand{\arraystretch}{1.15}

\begin{tabularx}{\textwidth}{
    l
    l
    c
    c
    c
    c
    >{\raggedright\arraybackslash}X
}
\hline
\textbf{Match} &
\textbf{Event} &
\textbf{GPT-5.4} &
\textbf{Conf.} &
\textbf{FlowJudge} &
\textbf{GPT-5.6} &
\textbf{Explanation} \\
\hline

Match 1 &
Goal &
8.61 &
0.91 &
9.0 &
9.0 &
Ferran Torres scores after a cross, with the ball inside the goal and sustained celebration from the scorer and team. The commentator's disbelief at two goals in two minutes makes the moment highly exciting. \\

Match 2 &
Shot &
8.62 &
0.93 &
6.0 &
8.0 &
The goalkeeper saves a close-range Aguero attempt after a City move into the penalty area. The commentator's ``to seal it for City! What a save'' makes the moment highly dramatic. \\

Match 3 &
Miss &
5.83 &
0.88 &
4.0 &
6.0 &
Bruno Fernandes spots Ortega off his line and attempts a long-range effort toward an open goal, but it goes wide. The near-open-goal setup adds tension, while the miss and muted commentary limit the excitement. \\

\hline
\end{tabularx}
\end{table*}
\section{Conclusion}

This work presented a zero-shot multimodal framework for soccer event spotting,
temporal localization, and excitement scoring from complete match broadcasts
without task-specific parameter fine-tuning. Across three annotated matches
with 130 events from six categories, we evaluated multiple models, input
representations, temporal settings, and prompt-optimization strategies. Combining
visual and commentary evidence improved event spotting, while storyboard input
provided strong performance at lower inference cost than individual-frame
multimodal input. Prompt optimization changed precision--recall and class-level
behavior but did not consistently improve overall performance.

Excitement scores showed positive rank agreement with two LLM-based evaluators, with particularly strong agreement from the GPT-based judge. However, the excitement analysis remains model-based and requires human validation to establish alignment with perceived highlight relevance. The three-match evaluation limits conclusions about generalization across leagues, broadcast styles, and commentary conditions. Sparse frame sampling may omit short-duration visual cues, while replay and recap sequences can introduce false-positive detections. CALF and NetVLAD++ are evaluated cross-dataset without retraining, so annotation and broadcast differences may affect the comparison.


Overall, the results demonstrate the potential of general-purpose multimodal
models for soccer video understanding, while motivating larger-scale evaluation,
better replay handling, and human validation of highlight relevance.


\begin{thebibliography}{00}

\bibitem{b1}
S. Giancola, M. Amine, T. Dghaily, and B. Ghanem,
``SoccerNet: A scalable dataset for action spotting in soccer videos,''
in \textit{Proc. IEEE/CVF Conf. Comput. Vis. Pattern Recognit. Workshops (CVPRW)},
2018, pp. 1711--1721.

\bibitem{b2}
A. Deliege, A. Cioppa, S. Giancola, M. J. Seikavandi,
J. V. Dueholm, K. Nasrollahi, B. Ghanem, T. B. Moeslund,
and M. Van Droogenbroeck,
``SoccerNet-v2: A dataset and benchmarks for holistic understanding of broadcast soccer videos,''
in \textit{Proc. IEEE/CVF Conf. Comput. Vis. Pattern Recognit. Workshops (CVPRW)},
2021, pp. 4503--4514.

\bibitem{b3}
A. Cioppa, A. Deliege, S. Giancola, B. Ghanem,
M. Van Droogenbroeck, R. Gade, and T. B. Moeslund,
``A context-aware loss function for action spotting in soccer videos,''
in \textit{Proc. IEEE/CVF Conf. Comput. Vis. Pattern Recognit. (CVPR)},
2020, pp. 13126--13136.

\bibitem{b4}
B. Vanderplaetse and S. Dupont,
``Improved soccer action spotting using both audio and video streams,''
in \textit{Proc. IEEE/CVF Conf. Comput. Vis. Pattern Recognit. Workshops (CVPRW)},
2020, pp. 3921--3931.







\bibitem{b5}
J. Rao, H. Wu, H. Jiang, Y. Zhang, Y. Wang, and W. Xie,
``Towards universal soccer video understanding,''
in \textit{Proc. IEEE/CVF Conf. Comput. Vis. Pattern Recognit. (CVPR)},
2025, pp. 8384--8394.

\bibitem{b6}
H. Yang, J. Rao, H. Wu, and W. Xie,
``SoccerMaster: A vision foundation model for soccer understanding,''
in \textit{Proc. IEEE/CVF Conf. Comput. Vis. Pattern Recognit. (CVPR)},
2026, pp. 21549--21560.

\bibitem{b7}
S. Ren, L. Yao, S. Li, X. Sun, and L. Hou,
``TimeChat: A time-sensitive multimodal large language model for long video understanding,''
in \textit{Proc. IEEE/CVF Conf. Comput. Vis. Pattern Recognit. (CVPR)},
2024, pp. 14313--14323.

\bibitem{b8}
L. A. Agrawal, S. Tan, D. Soylu, N. Ziems, R. Khare,
K. Opsahl-Ong, A. Singhvi, H. Shandilya, M. J. Ryan,
M. Jiang, C. Potts, K. Sen, A. G. Dimakis, I. Stoica,
D. Klein, M. Zaharia, and O. Khattab,
``GEPA: Reflective prompt evolution can outperform reinforcement learning,''
2025, arXiv:2507.19457.

\bibitem{b9}
K. Opsahl-Ong, M. J. Ryan, J. Purtell, D. Broman,
C. Potts, M. Zaharia, and O. Khattab,
``Optimizing instructions and demonstrations for multi-stage language model programs,''
in \textit{Proc. 2024 Conf. Empirical Methods Natural Language Processing (EMNLP)},
2024, pp. 9340--9366.

\bibitem{b10}
A. Xarles, S. Escalera, T. B. Moeslund, and A. Clap\'es,
``ASTRA: An action spotting TRAnsformer for soccer videos,''
in \textit{Proc. 6th Int. Workshop Multimedia Content Analysis Sports (MMSports)},
2023, pp. 93--102.

\bibitem{b11}
S. Giancola and B. Ghanem,
``Temporally-aware feature pooling for action spotting in soccer broadcasts,''
in \textit{Proc. IEEE/CVF Conf. Comput. Vis. Pattern Recognit. Workshops (CVPRW)},
2021, pp. 4490--4499.

\bibitem{b12}
A. Xarles, S. Escalera, T. B. Moeslund, and A. Clap\'es,
``T-DEED: Temporal-discriminability enhancer encoder-decoder for precise event spotting in sports videos,''
in \textit{Proc. IEEE/CVF Conf. Comput. Vis. Pattern Recognit. Workshops (CVPRW)},
2024, pp. 3410--3419.

\bibitem{b13}
J. Held, H. Itani, A. Cioppa, S. Giancola, B. Ghanem,
and M. Van Droogenbroeck,
``X-VARS: Introducing explainability in football refereeing with multi-modal large language models,''
in \textit{Proc. IEEE/CVF Conf. Comput. Vis. Pattern Recognit. Workshops (CVPRW)},
2024, pp. 3267--3279.

\bibitem{b14}
T. Jiang, H. Wang, M. S. Salekin, P. Atighehchian, and S. Zhang,
``Domain adaptation of VLM for soccer video understanding,''
in \textit{Proc. IEEE/CVF Conf. Comput. Vis. Pattern Recognit. Workshops (CVPRW)},
2025, pp. 6157--6167.

\bibitem{b15}
H. Wang, Z. Xu, Y. Cheng, S. Diao, Y. Zhou, Y. Cao,
Q. Wang, W. Ge, and L. Huang,
``Grounded-VideoLLM: Sharpening fine-grained temporal grounding in video large language models,''
2024, arXiv:2410.03290.

\bibitem{b16}
X. Zeng, K. Li, C. Wang, X. Li, T. Jiang, Z. Yan,
S. Li, Y. Shi, Z. Yue, Y. Wang, Y. Wang, Y. Qiao, and L. Wang,
``TimeSuite: Improving MLLMs for long video understanding via grounded tuning,''
2024, arXiv:2410.19702.

\bibitem{b17}
T. Kumakura, R. Orihara, Y. Tahara, A. Ohsuga, and Y. Sei,
``ASPERA: Multimodal soccer action spotting with visual, auditory, and textual modalities,''
\textit{IEICE Trans. Inf. Syst.},
vol. E109-D, no. 8, pp. 1205--1218, 2026.

\bibitem{b18}
S. Santra, V. Chudasama, P. Wasnik, and V. N. Balasubramanian,
``Precise event spotting in sports videos: Solving long-range dependency and class imbalance,''
in \textit{Proc. IEEE/CVF Conf. Comput. Vis. Pattern Recognit. (CVPR)},
2025, pp. 3163--3172.

\bibitem{b19}
C. Spearman,
``The proof and measurement of association between two things,''
\textit{The American Journal of Psychology},
vol. 15, no. 1, pp. 72--101, 1904.

\bibitem{b20}
Flow AI,
``Flow Judge: An open small language model for LLM system evaluations,''
2024. [Online]. \url{https://flow-ai.com/blog/flow-judge}

\end{thebibliography}
\end{document}